\documentclass{article}
\usepackage{tech2025}

\usepackage{microtype}
\usepackage{graphicx}
\usepackage{subfigure}
\usepackage{booktabs} 

\usepackage{hyperref}

\usepackage[utf8]{inputenc} 
\usepackage[T1]{fontenc}    
\usepackage{url}            
\usepackage{amsfonts}       
\usepackage{nicefrac}       
\usepackage[table]{xcolor}         
\usepackage{colortbl}
\usepackage{amsmath}
\usepackage{amssymb}
\usepackage{mathtools}
\usepackage{amsthm}
\usepackage{mathrsfs}
\usepackage{xspace}
\usepackage{academicons}

\usepackage[capitalize,noabbrev]{cleveref}

\theoremstyle{plain}

\theoremstyle{definition}

\theoremstyle{remark}

\theoremstyle{plain}
\newtheorem*{theorem*}{Theorem}
\newtheorem*{proposition*}{Proposition}
\newtheorem*{lemma*}{Lemma}
\newtheorem*{corollary*}{Corollary}

\theoremstyle{definition}
\newtheorem*{definition*}{Definition}
\newtheorem*{assumption*}{Assumption}

\theoremstyle{remark}
\newtheorem*{remark*}{Remark}

\definecolor{bigaired}{RGB}{156, 0, 0}
\definecolor{uclablue}{RGB}{39, 116, 174}
\definecolor{thupurple}{RGB}{102, 8, 116}
\definecolor{pkured}{RGB}{139, 0, 18}
\definecolor{panton}{RGB}{217, 51, 121}

\definecolor{darkred}{RGB}{200, 0, 0}
\definecolor{darkblue}{RGB}{0, 0, 200}
\definecolor{blue}{RGB}{0, 0, 200}

\definecolor{light}{RGB}{225, 250, 250}
\definecolor{lightgray}{RGB}{0.9, 0.9, 0.9}
\definecolor{lightred}{RGB}{250, 200, 200}
\definecolor{lightblue}{RGB}{210, 220, 250}
\definecolor{lightpurple}{RGB}{218,210,255}

\definecolor{doderblue}{RGB}{30, 144, 255}
\definecolor{select}{RGB}{222, 235, 247}
\definecolor{unselect}{RGB}{247, 207, 206}

\definecolor{myLinkColor}{RGB}{0, 0, 200}     
\definecolor{myCiteColor}{RGB}{0, 0, 200}     
\definecolor{myURLColor}{RGB}{0, 0, 200}      
\hypersetup{
  colorlinks=true,
  linkcolor=myLinkColor,
  citecolor=myCiteColor,
  urlcolor=myURLColor
}

\usepackage[textsize=tiny]{todonotes}

\usepackage[most]{tcolorbox}

\usepackage{listings}

\usepackage{fontawesome5}  
\usepackage{listings}      
\usepackage[misc]{ifsym}
\usepackage{wrapfig}    
\usepackage[T1]{fontenc}

\usepackage{tcolorbox}
\usepackage{relsize}

\usepackage{adjustbox}
\usepackage{fancyhdr}
\usepackage{lipsum}
\usepackage{newtxtext}
\usepackage{wrapfig}
\usepackage{enumitem}
\definecolor{azblue}{RGB}{27,117,187}      

\usepackage[noend]{algpseudocode}   

\definecolor{bestcol}{RGB}{  0,102,204} 
\definecolor{goodcol}{RGB}{ 34,139, 34} 
\definecolor{deltaBg}{RGB}{220,230,255} 

\usepackage{lmodern}  

\usepackage{tikz}
\usetikzlibrary{tikzmark,decorations.pathreplacing,calc}
\usepackage{stackengine}
\stackMath
\definecolor{lightgreen}{RGB}{0,150,0}  

\usepackage{titletoc}

\newtheoremstyle{rqstyle}%
  {\topsep}            
  {\topsep}            
  {}                   
  {}                   
  {\bfseries}    
  {:}                  
  {.5em}               
  {}                   

\theoremstyle{rqstyle}

\crefname{researchquestion}{Research Question}{Research Questions}

\definecolor{propose}{HTML}{EF8E8D}
\definecolor{solve}{HTML}{5755A3}

\definecolor{humanred}{RGB}{180, 50, 50}
\definecolor{envgreen}{RGB}{50, 140, 80}

\definecolor{paleviolet}{HTML}{E1EEFC}
\definecolor{lightgrey}{RGB}{247, 247, 247}
\newenvironment{leapabstract}{
  \begin{tcolorbox}[
    colback=lightgrey,
    colframe=white,
    boxrule=0pt,
    arc=10pt,
    left=16pt,
    right=16pt,
    top=12pt,
    bottom=12pt,
    width=\textwidth,
    enlarge left by=0mm,
    before skip=10pt,
    after skip=10pt
  ]
  \normalsize
}{
  \end{tcolorbox}
}

\makeatletter
\DeclareRobustCommand\onedot{\futurelet\@let@token\@onedot}
\def\@onedot{\ifx\@let@token.\else.\null\fi\xspace}

\makeatother

\usepackage{graphicx}%
\usepackage{multirow}%
\usepackage{amsmath,amssymb,amsfonts}%
\usepackage{amsthm}%
\usepackage{mathrsfs}%
\usepackage[figuresright]{rotating}%
\usepackage[title]{appendix}%
\usepackage{xcolor}%
\usepackage{textcomp}%
\usepackage{manyfoot}%
\usepackage{booktabs}%
\usepackage{algorithm}%
\usepackage{algorithmicx}%
\usepackage{algpseudocode}%
\usepackage{program}%
\usepackage{listings}%

\usepackage{color}
\usepackage{float}
\usepackage{etoolbox}
\usepackage{bm}
\usepackage{enumitem}
\usepackage{pdfpages}
\usepackage{lineno}

\begin{document}

\makeatletter
\def\icmldate#1{\gdef\@icmldate{#1}}
\icmldate{\today}
\makeatother

\makeatother


\icmltitlerunning{
    SCOUT: Fast Spectral CT Imaging in Ultra LOw-data Regimes via PseUdo-label GeneraTion
}
\vskip 0.05in
\icmltitle{
     SCOUT: Fast Spectral CT Imaging in Ultra LOw-data Regimes via PseUdo-label GeneraTion
}


\begin{icmlauthorlist}
\mbox{
  Guoquan Wei$^{1}$
  }
\mbox{
  Liu Shi\ \!$^{1,\ \textrm{\Letter}}$
  }
\mbox{
  Shaoyu Wang$^{1}$
  }
\mbox{
  Mohan Li$^{2,3,4}$
  }
\mbox{
  Cunfeng Wei$^{2,3,4}$
  }
\mbox{
  Qiegen Liu\ \!$^{1,\ \textrm{\Letter}}$
  }
\end{icmlauthorlist}
\\
{
  \small
   $^{1}$School of Information Engineering, Nanchang University, Nanchang, 330031, Jiangxi, China.\\
   $^{2}$Institute of High Energy Physics, Chinese Academy of Sciences, Beijing, 100049, China.\\
   $^{3}$Jinan Laboratory of Applied Nuclear Science, Jinan, 251401, Shandong, China.\\
   $^{4}$Ray Image Testing Technology (Jinan) Co. Ltd, Jinan, Shandong, China.\\
  ${\textrm{\Letter}}$ Corresponding Authors: Liu Shi, Qiegen Liu.\\
  \texttt{\{guoquanwei\}@email.ncu.edu.cn, \{shiliu, liuqiegen\}@ncu.edu.cn}.
}




\begin{leapabstract}
    Noise and artifacts during computed tomography (CT) scans are a fundamental challenge affecting disease diagnosis. However, current methods either involve excessively long reconstruction times or rely on data-driven models for optimization, failing to adequately consider the valuable information inherent in the data itself, especially medical 3D data. This work proposes a reconstruction method under ultra-low raw data conditions, requiring no external data and avoiding lengthy pre-training processes. By leveraging spatial nonlocal similarity and the conjugate properties of the projection domain to generate pseudo-3D data for self-supervised training, high-fidelity results can be achieved in a very short time. Extensive experiments demonstrate that this method not only mitigates detector-induced ring artifacts but also exhibits unprecedented capabilities in detail recovery. This method provides a new paradigm for research using unlabeled raw projection data.
    Code is available at \url{https://github.com/yqx7150/SCOUT}.

    
\end{leapabstract}


\vskip -0.1in
\section{Introduction}
\label{sec:introduction}

Photon-counting computed tomography (PCCT) is a more advanced imaging method due to its use of photon-counting detectors with smaller pixel sizes \cite{willemink2018photon,rajendran2022first,du2025x}. Compared to conventional computed tomography (CT), its high resolution and multi-energy-segment data acquisition give it significant advantages in specialized fields such as vascular, bone and soft tissue imaging, and material decomposition \cite{douek2023clinical,sharma2024photon}. However, a narrower energy window leads to a reduction in the number of incident photons, resulting in increased noise in the reconstructed image. At the same time, inconsistent detector response and damage can cause global ring artifacts in the image \cite{saleh2005multiplied,hasinoff2021photon}. Therefore, researching simple and efficient optimization methods is crucial for clinical applications.

 Over the past few decades, numerous researchers have explored the local or global prior characteristics of PCCT multi-channel data \cite{wu2019block,guo2023spectral2spectral,xie2016multispectral,wu2018non}. However, these methods are tailored for specific scenarios and are not adaptable to data acquired through different imaging modalities. Moreover, the massive scale and computational complexity of processing multi-channel data present formidable challenges. For instance, PCCT simultaneously acquires data across multiple energy bands, with raw projection data from a single scan easily reaching tens of gigabytes per channel. This exponential growth in spatial and temporal dimensions imposes stringent demands on computational hardware, resulting in substantial memory consumption and excessive processing times. Even in studies focusing on a single energy channel, the multi-parameter nature, empirical reliance, and time-consuming nature of existing methods compromise robustness, limiting their broader application \cite{xu2012image,zhao2012dual}.

In recent years, the emergence of deep learning has brought great convenience to improving the imaging effect of medical images and enhancing diagnostic results \cite{rajpurkar2023current,shan2019competitive,wang2020deep,shen2022mlf}. However, current research focuses too much on improving model performance and rarely considers the characteristics of CT data, such as the conjugate theorem in the projection domain \cite{shi2025zero}. In clinical scenarios, it is difficult to collect paired data, and the end-to-end supervised form limits its generalization ability \cite{chen2017low,huang2021gan,wu2021drone,zhang2021clear}. Large language foundation models \cite{wang2025self,wu2025towards,wu2025vision,rao2025multimodal,qiu2024towards,he2024foundation} are supported by massive amounts of data and can achieve general or special tasks, but the powerful computing power required limits clinical applications. At the same time, the emergence of diffusion models \cite{croitoru2023diffusion,yang2023diffusion,ho2020denoising,song2020denoising,chung2022improving,song2020score} has also made the denoising task very popular. However, this type of method has problems such as redundant sampling steps and excessively long training time. More importantly, most current research focuses on studying two-dimensional imaging relationships, while ignoring the dependencies, similarities and other characteristics that exist in space \cite{gao2023corediff}. We have noticed that self-supervised recovery, which does not require labels and relies solely on its own data, is very suitable for clinical scenarios \cite{lehtinen2018noise2noise,zeng2022accurate,li2024self,lu2025physics,li2023spatial,Guo_2026}. For PCCT, current methods are mainly divided into non-local self-similarity, global spectral channel use, and combined use \cite{niu2022noise,shi2025zs4d,lin2024loquat}, with very few studies on the physical characteristics of the original data \cite{tulin2025self,gao2025noise,unal2024proj2proj}. Even some current self-supervised research methods still rely on pre-training with defective external data before testing. This fitting method inevitably loses high-frequency information, resulting in a smooth imaging effect.

This work proposes a zero-shot method for pseudo-label generation to address these challenges. Specifically, inspired by the similarity between adjacent slices in medical data and the principles of projection, we shift our focus to raw 3D projective data. On one hand, we utilize spatial nonlocal similarity to find multiple voxels similar to the current voxel and then distribute them in parallel across multiple new datasets to form a low-rank 3D voxel bank. On the other hand, to prevent imaging illusions caused by statistics, we introduce the conjugation properties of projection into the spatial permutation level, rapidly generating another batch of physically constrained voxel bank. Notably, our proposed method exclusively requires a single 3D low-dose projection volume, operating completely free from any reliance on external datasets or even correlated multi-channel information. This remarkable efficiency enables the comprehensive processing of a volumetric dataset containing approximately $300-600$ human slices in a mere $3$ to $10$ minutes. Such ultra-fast imaging speed makes it entirely feasible to process massive multi-channel data independently, effectively bypassing the traditional computational bottlenecks of spectral CT processing. Furthermore, the inherent versatility of this framework extends its potential to conventional CT and various other sampling modalities. Extensive experiments demonstrate that our approach comprehensively characterizes and mitigates complex noise distributions and artifact defects, thereby bringing exceptional convenience and robust clinical applicability to both PCCT mouse imaging and human abdominal or pulmonary diagnostics.

\begin{figure*}[!t]
    \centering
    \includegraphics[width=\textwidth]{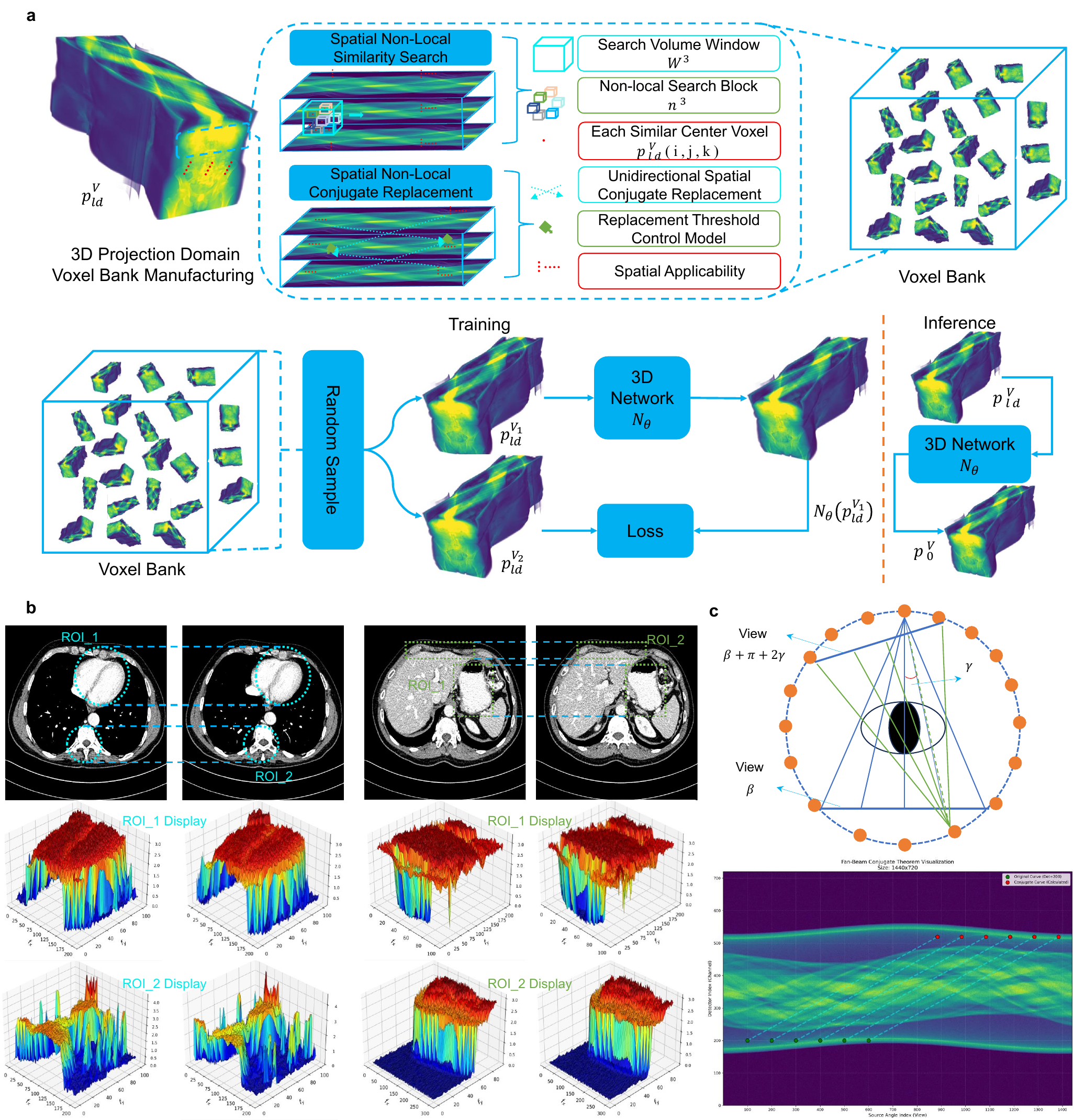}
    \caption{
        \textbf{Schematic diagram of the overall workflow and principles of SCOUT.}
        \textbf{(a) \textit{(top half)}} Introduces the detailed process of creating a voxel library starting from low-dose CT projection data, using spatial nonlocal similarity and the conjugate theorem.
        \textbf{(a) \textit{(bottom half)}} After obtaining the voxel library, two original volume projection data are randomly selected for self-supervised training, followed by testing to obtain denoised volume data.
        \textbf{(b)} Introduces the similar anatomical structures that exist in large numbers in human data, laying the groundwork for explaining the existence of the same similarity in the projection domain.
        \textbf{(c)} Intuitively demonstrates the conjugate symmetry caused by the scanning process, which also supports the creation of pseudo-label data. Then, it presents the manifestation of conjugate properties in single slices, providing theoretical support for extending to domain slices.
        \label{fig1}
        }
    \vskip 0.2in
\end{figure*}

\section{Results}

In this section, we first briefly introduce the concept and process of SCOUT, then introduce the effectiveness of removing the inherent noise generated by the acquisition equipment when the tube current and exposure time are sufficient, then introduce the excellent reproducibility of dual-energy walnut data in low-dose scenarios, and finally introduce the universality and versatility of traditional CT in low-dose, ultra-low-dose and ring artifact data. More details can be
found in Section \ref{sec:method}.

\subsection{Overview of SCOUT}
\label{Overview of SCOUT}

X-ray computed tomography is essentially a series of photon statistics captured along a specific trajectory. Regardless of whether a circular or helical trajectory is used in parallel, fan-beam, or cone-beam geometry, the detector records the attenuation coefficients of objects from multiple angles. This data, presented as a sine wave, contains the fundamental physical information that any algorithm reconstructs to the image domain. Compared to the complex noise correlations and artifacts in the image domain, the quantum noise in the projection domain data exhibits a Poisson distribution, preserving the original signal-to-noise ratio and making it an ideal processing domain for improving image quality. However, individual projection data and their corresponding images are often processed separately. To address this challenge, we propose SCOUT, a zero-shot self-supervised fast imaging method for 3D projection domain data. This method fully utilizes the widespread similarity of the original volume data to generate pseudo-labels, and its overall process is shown in Fig. \ref{fig1}a. Specifically, this is achieved by synergistically leveraging two key characteristics. First, we reveal the abundance of similar information within the projection data. Numerous anatomical structures, including bones, soft tissues, and lesions, frequently recur throughout the volumetric space, as shown in Fig. \ref{fig1}b. Unlike previous studies that focused on local areas, we enhance information utilization by broadly searching for similar information globally. For any given region in the volumetric data, there exists a twin that is extremely similar to the currently selected region. By aggregating these similar data, we can expand the data volume from a single instance to a statistical set. Second, beyond this statistical similarity, the acquisition process itself is subject to a strict geometric constraint. Taking parallel beam geometry as an example, the scanning trajectory induces a deterministic conjugate symmetry: the conjugate acquisition point corresponding to the point $p(s,\theta)$ incident on detector $s$ at an angle $\theta$ is $p(-s,\theta+\pi)$, as illustrated in Fig. \ref{fig1}c. Utilizing both statistical similarity and geometric properties, we can achieve pseudo-label generation, enabling the self-supervised network to process entirely based on the original data itself. The low-rank nature of the information distribution and the high-rank nature of the noise promote a good separation between the two.

\subsection{Applying SCOUT to Address Intrinsic Noise in Data Acquisition Equipment}
\label{Applying SCOUT to Address Intrinsic Noise in Data Acquisition Equipment}
\begin{figure*}[!t]
	\centering
		\includegraphics[width=\textwidth]{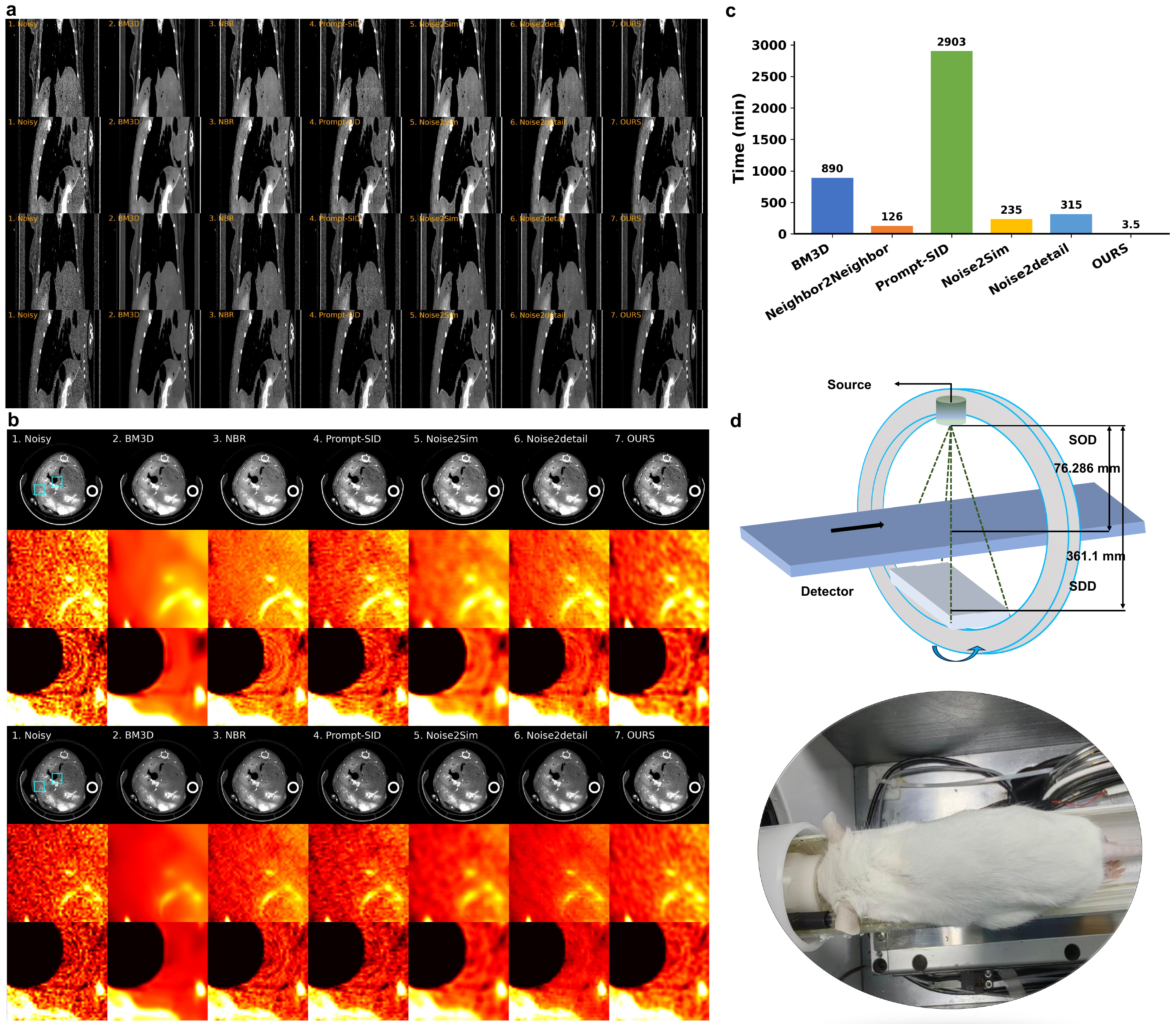}
	\caption{
		\textbf{Mouse and acquisition equipment model and results description.}
		\textbf{(a)} The coronal and sagittal plots of the self-supervised method with good results.
		\textbf{(b)} Axial slice results showing that the second and third rows are magnified views of the selected regions of interest.
		\textbf{(c)} The record shows the total time spent on training and testing. This demonstrates that our algorithm remains extremely fast even when processing volumetric data.
		\textbf{(d) \textit{(top half)}} The scanning geometry of the mouse acquisition system $\mu$Color SA. Right half, the mice used in the experiment.
		\textbf{(d) \textit{(bottom half)}} The mice used in the experiment.
		\label{fig2}
	}
	\vskip 0.2in
\end{figure*}

Due to the tiny size of animals like mice, high-resolution detectors are needed to capture intricate microstructures. Simultaneously, acquiring information across different energy ranges requires energy segmentation of X-rays. Furthermore, to obtain spectral information, the detector must further segment the limited X-ray flux along the energy dimension. Spatial voxels and narrow energy windows lead to insufficient incident photons and noise perturbations. In addition, the inherent pixel-to-pixel response inconsistencies in photon-counting detectors manifest as structured ring artifacts in the image domain, affecting the accurate material decomposition process. Despite these challenges, PCCT, capable of accurately distinguishing bone structures, iodine-contrast-perfused soft tissue, and calcified lesions through material decomposition without multiple scans, remains a cutting-edge technology driving the development of tumor angiography and pharmacokinetics. However, the imaging limitations inherent in the aforementioned equipment scanning methods still hinder further progress in life science research. To better advance PCCT, our few-sample fast imaging algorithm can achieve real-time improvement of imaging results in the projection domain. As shown in Fig. \ref{fig2}a and b, the data from the first two low-energy channels demonstrate that our method not only improves noise and reduces ring artifacts but also preserves detail. The mice were scanned after contrast agent injection, and the visual results show that our method makes the distribution of the contrast agent in each channel clearer. Compared with Noise2sim \cite{niu2022noise}, our method shows advantages in desharpening and detail preservation. While other comparative methods also demonstrated effectiveness compared to the original data, they were less effective in achieving good imaging results.

\begin{figure*}[!t]
	\centering
		\includegraphics[width=\textwidth]{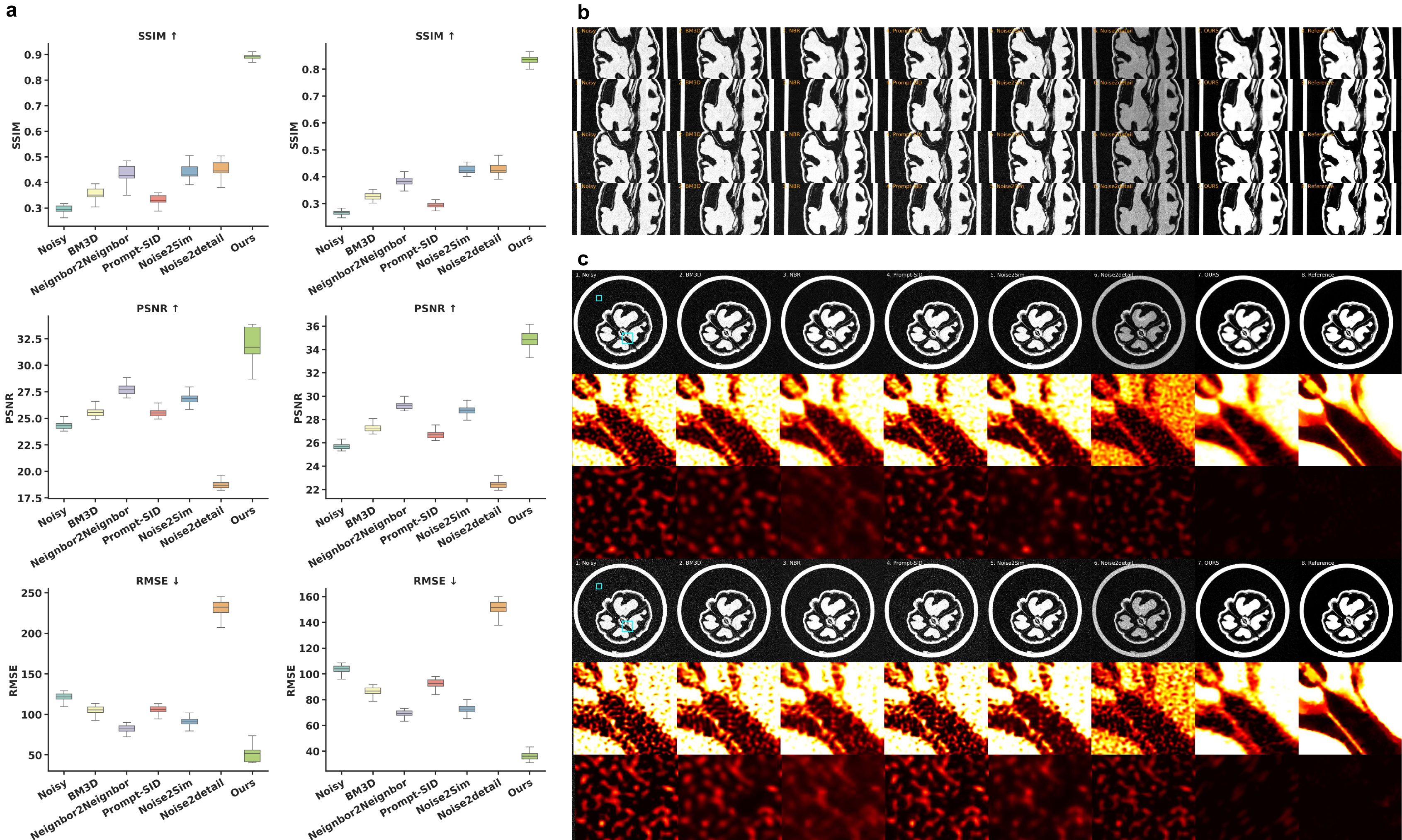}
	\caption{
		\textbf{Relevant results from the walnut data.}
		\textbf{(a)} Box plots of PSNR, SSIM, and RMSE for all comparative experiments and reference reconstructed images.
		\textbf{(b)} Coronal and sagittal plots.
		\textbf{(c)} Axial plane dual-energy spectral data results and magnified views of the region of interest.
		\label{fig3}
	}
	\vskip 0.2in
\end{figure*}

Besides ensuring imaging quality, another crucial aspect of CT scans is reconstruction time. The advent of deep learning has made fast and efficient imaging possible; however, current self-supervised methods primarily focus on 2D slice data, lacking spatial awareness and consuming significant computational resources for the iterative processing of numerous slices. Furthermore, post-processing reconstructed image data cannot truly resolve noise, artifacts, and sparsity issues that originate in the projection domain. By directly targeting 3D raw projection data, our method not only addresses these physical degradations at their source but also drastically reduces optimization time. Since PCCT mouse scans utilize axial continuous scanning with segmented data distribution, our algorithm can refine the previous segment while the next is being acquired, achieving near real-time optimization directly in the raw projection domain. As quantitatively demonstrated in Fig. \ref{fig2}c, our approach achieves an unprecedented total training and testing time of merely 3.5 minutes. This represents a staggering acceleration compared to existing methods: it is exactly 36 times faster than the highly competitive Neighbor2Neighbor method at 126 minutes, and delivers an astonishing speedup of over 820 times compared to Prompt-SID at 2903 minutes. Even against other efficient networks like Noise2Sim taking 235 minutes and Noise2detail taking 315 minutes, our method reduces the processing time by approximately two orders of magnitude, firmly establishing its capability for ultra-fast, high-throughput imaging scenarios.

\subsection{The Remarkable Performance in Walnut Dual-Energy CT Data}
\label{The Remarkable Performance in Walnut Dual-Energy CT Data}
To fully demonstrate the algorithm's effectiveness on heterogeneous data, we selected publicly available walnut data \cite{zhou2025cone} for further research. Walnuts, like mice, possess unique anatomical structures that are highly valued by researchers. The shell, the soft and varied-shaped flesh, and the internal air cavities make the data characteristics resemble the head information distribution of small animals, making it a valuable resource for analyzing the algorithm's effectiveness in noise reduction. We also processed the high and low energy ranges of this data separately, simulating noise maps across different channels based on the photon count and scanning parameters published in the original paper. Fig. \ref{fig3}a shows that the algorithm's PSNR, SSIM, and RMSE are significantly higher than the noise map and the comparison method, while also demonstrating excellent results in the coronal, sagittal, axial planes, and magnified regions of interest as shown in Fig. \ref{fig3}b and c. For some persistent noise that remains, we believe the algorithm can recover details to some extent through iterative processing and dual-domain processing. For tissues completely destroyed by noise, complete recovery without the aid of real images remains a challenge. In summary, different incident photon numbers between high and low energy bands cause different noise distributions, but this method can still recover them to a large extent. This fully demonstrates that our method is not limited by the shape and size of the object and has strong versatility and clinical applicability.

\begin{figure*}[!t]
	\centering
		\includegraphics[width=\textwidth]{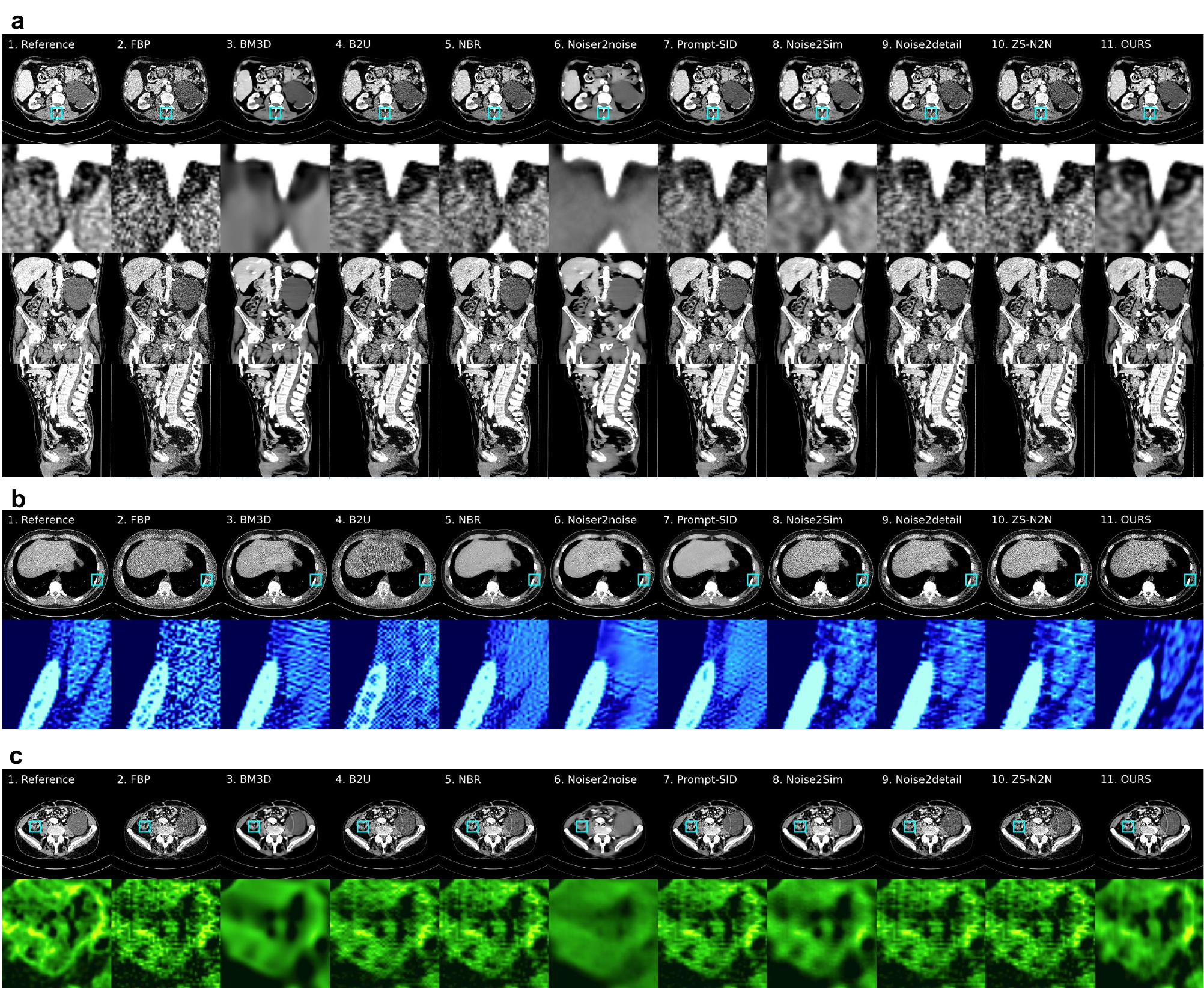}
	\caption{
		\textbf{Results related to traditional medical human data.}
		\textbf{(a)} Quantitative evaluation results on Mayo 2016 data, with the first and second rows showing axial slices and magnified views of the region of interest. The third and fourth rows show the coronal and sagittal results of the volume data.
		\textbf{(b)} Results demonstrating effectiveness at ultra-low doses.
		\textbf{(c)} Experimental results addressing ring artifacts involving bright streaks, dark streaks, and broad streaks.
		\label{fig4}
	}
	\vskip 0.2in
\end{figure*}

\subsection{Universality of Data from Traditional Energy Integral Detectors}
\label{Universality of Data from Traditional Energy Integral Detectors}
Although our method is designed for PCCT, the advantage of processing individual volume data is the universality of the algorithm's application. Even for PCCT, we process single-channel data, a rationale driven by the rapid imaging process. While traditional CT lacks energy partitioning, each channel in PCCT, like its predecessor, has an attenuation coefficient. Therefore, we conducted extensive experiments on four major public datasets—Mayo2016 \cite{chen2016open}, Mayo2020 \cite{moen2021low}, LIDC-IDRI \cite{armato2011lung}, and CTspine1K \cite{deng2021ctspine1k}—covering complex anatomical variations from the chest and abdomen to the spine, to validate the method's versatility in improving image quality. We simulated sine waves with planar fan-beam properties using the Torch-radon \cite{ronchetti2020torchradon} toolkit and simulated real degradation processes by injecting mixed Gaussian-Poisson noise. As can be seen from Fig. \ref{fig4}a, our method exhibits the best detail recovery in transverse, coronal, and sagittal planes. Quantitative evaluation shows that our method improves PSNR by an average of 3 dB and SSIM by 7\%, establishing a significant statistical advantage. More importantly, we have overcome the computational complexity barrier of traditional iterative algorithms such as BM3D \cite{dabov2006image}. Compared to the computational difficulties of traditional nonlocal methods that require hours to process full-body data, our method can complete the entire reconstruction process in minutes. This leap in speed from offline processing to online inference highlights its deployment potential in the time-sensitive clinical environment.

Furthermore, to investigate the robustness of the algorithm under ultra-low incident photon conditions, we directly utilized low-dose data provided by Mayo 2020. As shown in Fig. \ref{fig4}b, under extremely low signal-to-noise ratio conditions, the comparison method either fails to distinguish between signal and noise, resulting in noise residue or even amplification. Similarly, we simulated ring artifacts under low-dose conditions to verify the ability to reduce ring artifacts. As shown in Fig. \ref{fig4}c, although Noise2sim \cite{niu2022noise} improves noise and reduces artifacts, this comes at the cost of sacrificing detail and smoothing key features. However, medical images are geared towards medical diagnosis, and detail is crucial for improving the accuracy of doctors' diagnoses. Both of these experiments demonstrate the effectiveness of our method in terms of both image quality and evaluation metrics. It is worth emphasizing that all experiments show the training and testing times, demonstrating that our method achieves the advantages of simple algorithm, fast imaging speed, and high image quality.

\section{Discussion}
\label{sec:Discussion}
This work aims to propose a new paradigm for rapid imaging based on raw 3D projection data without requiring any external data. Since the purpose of medical imaging is to present the patient's internal condition and help doctors improve diagnostic accuracy, previous studies utilizing data partitioning, data blind spots, and channel correlations have relied heavily on pre-training and fitting, which can lead to illusions and pose a risk to medical judgment. Our method for generating pseudo-samples employs a global search, and conjugate physical constraints ensure high similarity between pseudo-samples, ideally resembling a rank-1 matrix. Furthermore, relying solely on independent 3D data as a starting point allows it to be applied to any scanning mode, scanning geometry, and heterogeneous data. Similarly, our method is built upon raw projection data, enabling us to better perceive the distribution of anatomical information. Moreover, the algorithm is simple to implement, fast in processing, and lightweight, making it a highly promising technology.

While this method demonstrates good robustness and applicability, some aspects warrant attention and improvement. First, its fundamental assumption—that adjacent pixels should have similar structures—should be carefully considered. If the spatial sampling rate is too low, insufficient redundancy will be provided, leading to reduced imaging quality. Furthermore, since 3D data itself has a certain amount of data volume, generating additional samples requires the processing device to have sufficient storage capacity. The network used in this method is only a few simple convolutional layers. We believe that there are many more excellent networks that can be applied to perceive more useful information, such as U-Net and Transformer, but for the sake of lightweight design and speed, they have not been discussed in detail. We also emphasize that although our projection domain research mainly uses fan-beam flat panel detectors, the data for the mice and walnuts were acquired using cone-beam CT. The acquired projection data, when arranged in sequence, will still yield projection volume data, so this method has strong universality.

\section{Methods}
\label{sec:method}
This section introduces the equipment used for mouse collection, the detailed pseudo-label generation process, and the training process. It then describes the application of similarity heuristics to the projection domain (spatial nonlocality and the conjugate theorem between neighborhood slices). Subsequent sections introduce detailed data simulation methods, comparison methods, and evaluation metrics.
\subsection{Experimental Equipment}
The PCCT mouse data were acquired using the $\mu$Color SA system developed by the Institute of High Energy Physics, Chinese Academy of Sciences. This device acquires $2000$ projection data frames of $2062 \times 252$ pixels each time. The scanning parameters were: source-to-object distance $76.286$ $mm$, source-to-detector distance $361.1$ $mm$, and pixel size $0.1$ $mm$. The device employs an advanced photon counting detector, achieving a spatial resolution of up to $15$ $\mu m$. It is primarily used for in vivo animal CT imaging in preclinical experimental studies, but can now also be used for whole-body structural energy spectrum imaging of various rodents. Specific scanning geometry is shown in Fig. \ref{fig2}d. We used this device to acquire dual-energy spectral CT data and four-channel data, with energy ranges of $21-30$ $keV$, $30-70$ $keV$, and $20-90$, $31-90$, $45-90$, $60-90$ $keV$, respectively. This experimental procedure complies with animal ethics guidelines and has been approved by the Ethics Committee of Nanchang University (Approval No.: 20220726008).

\subsection{Network Process and Objective Function}
The main workflow of this method consists of constructing a sample voxel bank, randomly sampling volumetric data for training, and subsequently conducting testing. Specifically, we first define $p_{ld}^{V_{0}}\in\mathbb{R}^{D\times H\times W}$ as the initial low-dose CT projection volumetric data, where $u\in\mathbb{R}$ represents the voxel value at spatial coordinates $(x,y,z)$. A 3D geometric block $p_{ld}^{V_{u}}\in\mathbb{R}^{n\times n\times n}$ is constructed centered at \(u\). Then, centered at $p_{ld}^{V_u}$, generate $k$ identical, height-similar volumetric blocks within a cube of size $W^3$. Subsequently, sort the $k$ similar non-local blocks according to similarity metrics based on Euclidean distances in space:

\begin{equation}
	P_u=
	\begin{Bmatrix}
		p_{ld}^{V_{u_1}},p_{ld}^{V_{u_2}},...,p_{ld}^{V_{u_k}}
	\end{Bmatrix}
	\;\label{eq1}
\end{equation}

where $u_k$ denotes the spatial position of the $k$-th non-local block. $P_u$ serves as a statistical repository containing samples, describing highly similar non-local blocks at different spatial positions. After a global traversal search, $k$ highly similar three-dimensional projection volumes form the sample bank $P_G$:

\begin{equation}
	P_G=\{p_{ld}^{V_1},p_{ld}^{V_2},\ldots,p_{ld}^{V_k}\}
	\label{eq2}	
\end{equation}

Furthermore, we leverage the conjugate property of the projection domain to perform spatial replacement at the slice level, as illustrated by Spatial Non-Local Conjugate Replacement in Fig. \ref{fig1}c. This can be described as follows: Select a proportion $\mathcal{P}_1$ of the volumetric data from $p_{ld}^{V_0}$, and based on this, select a proportion $\mathcal{P}_2$ of voxel values $u \in p\left(\beta,\gamma,z\right)$. Using conjugate symmetry as an operator, we search for voxel values $u_{c}\in p(\beta+\pi+2\gamma,-\gamma,z\pm m\Delta z)$ within the neighborhood $m\Delta z$. After matching, we use $u_c$ to replace $u$, generating a sample library $P_{G_c}$ of $k$ samples with physical geometric properties.

After obtaining the sample library, we expanded from a single sample to multiple samples. Due to the high similarity, the entire sample library exhibits low-rank characteristics, allowing us to better utilize neural networks for removal. According to the Noise2noise theory, the higher the similarity between two noisy samples entering the network, the closer the recovery effect is to the effect of clean signal supervision. Therefore, to avoid interference between the two sample libraries and to fully extract the distribution of clean information, we jointly trained the generated sample libraries. Specifically, there exists a $\lambda$ to control the extraction of elements from each sample library for training. In summary, the objective function of this method can be described as:

\begin{equation}\mathcal{L}_{total}=\lambda\cdot\mathbb{E}_{\left(p_{ld}^{V_{i}},p_{ld}^{V_{j}}\right)\sim{P_{G}}}\parallel N_{\theta}\left(p_{ld}^{V_{i}}\right)-p_{ld}^{V_{j}}\parallel_{2}^{2}+(1-\lambda)\cdot\mathbb{E}_{\left(p_{ld}^{V_{ic}},p_{ld}^{V_{jc}}\right)\sim{P_{Gc}}}\parallel N_{\theta}\left(p_{ld}^{V_{ic}}\right)-p_{ld}^{V_{jc}}\parallel_{2}^{2}
\label{eq3}
\end{equation}

where $N_\theta\left(\cdot\right)$ represents the recovery network. To fully utilize the spatial correlation in the data, this method processes all data in a 3D network, which mainly consists of five layers of 3D convolutional networks. Each layer is activated by a LeakyReLU function, and all convolutional kernels are $3\times3\times3$. Finally, a regular convolution is applied as the network output. Our network is very simple. Training is performed using the Adam optimizer with an initial learning rate of $0.001$, which decays every $500$ iterations, resulting in a total training iteration count of $3000-5000$. This allows us to process single-entity data in approximately $10$ minutes on a PyTorch-based 4090D (24GB) workbench.

\subsection{Explanation of Similarity of Projection Data}
Similarity is a crucial characteristic in medical data. It relies on the human body scanning process; the physical pixels of the detector determine the similarity between slices that are very close together. As shown in Fig. \ref{fig1}b, it can be seen that there are always some similar anatomical structures between images. Correspondingly, under the assumption of similarity in the image domain, based on the characteristics of CT scans and imaging principles, the original data domain possesses the same similarity property, as explained below: Define $f\left(x,y,z\right)$ as the attenuation value at $\left(x,y,z\right)$ in three-dimensional space, and $p\left(s,\theta,z\right)$ as the projection data acquired at detector position $s$, angle $\theta$, and located at axis $z$. According to the Radon transform, we know:

\begin{equation}p(s,\theta,z)=\mathcal{R}\{f(x,y,z)\}=\int_{-\infty}^{\infty}\int_{-\infty}^{\infty}f(x,y,z)\delta(x\mathrm{cos}\theta+y\mathrm{sin}\theta-s)dxdy
\label{eq4}
\end{equation}

where $\mathcal{R}(\cdot)$ is the Radon transform. The above formula allows for the orthographic projection of the image domain into the original data domain.

Since the images are similar, then:

\begin{equation}
	\Delta f=f(x,y,z+\Delta z)-f(x,y,z)
	\label{eq5}
\end{equation}

where $\Delta f$ represents the overall difference between slices, and due to the longitudinal continuity of human anatomy and the sufficiently small influence of detector pixels on $\Delta z$, $\|\Delta f\|$ is also a very small quantity to a certain extent, corresponding to a strong similarity between the two. According to Radon's linear properties, we have:

\begin{equation}
\begin{gathered}
		\Delta p=p(z+\Delta z)-p(z) \\
		=\mathcal{R}\{f(z+\Delta z)\}-\mathcal{R}\{f(z)\} \\
		=\mathcal{R}\{f(z+\Delta z)-f(z)\} \\
		=\mathcal{R}\{\Delta f\}
\end{gathered}
\label{eq6}
\end{equation}

since the scanned object is the human body, Eq. (\ref{eq4}) becomes:

\begin{equation}
	p(s,\theta,z)=\mathcal{R}\{f(x,y,z)\}=\int_{-2/H}^{2/H}\int_{-2/W}^{2/W}f(x,y,z)\delta(x\mathrm{cos}\theta+y\mathrm{sin}\theta-s)dxdy
	\label{eq7}
\end{equation}

then there must exist a constant $K$ such that $||\Delta p||=||\mathcal{R}\{\Delta f\}||\leq K\cdot||\Delta f||$, so the slices in the original string diagram still exhibit similarity.

\subsection{Data Operations and Simulation Instructions}
We provide a detailed description of the simulation methodology employed for the four major public datasets used in this work. First, the Mayo2016 dataset from the AAPM Grand Challenge contains data from $10$ patients. After acquiring reference images, we performed forward projection using the Torch-Radon tool and employed a simulated low-dose CT method similar to \cite{zeng2015simple}. This approach yielded data at various doses. The simulation employed an incident photon count of $2.5\times10^5$ and an electron noise variance of $10$. The scanning geometry parameters were: source-to-rotation center distance of $1361.2$ $mm$ and detector-to-rotation center distance of $615.18$ $mm$. A total of $1440$ views were acquired over a $360$-degree range, with a sinusoidal image size of $1440\times720$ pixels. Images were reconstructed using the fan-beam FBP algorithm at $512\times512$ resolution. The dataset simulated $25\%$ and $10\%$ dose levels, with $8$ cases used for training the comparison methods and $2$ cases ($L067$ and $L506$) for testing. Concurrently, we simulated various ring artifacts including bright rings, dark rings, and wide rings. The LIDC-IDRI dataset comprises 1,018 lung CT scans from $1,010$ lung cancer patients. We randomly selected 50 cases, performed forward projection, and sequentially stacked the results to generate 3D projection data. Mayo2020 contains $10$ cases with both low-dose and full-dose data; we directly utilized its low-dose data to validate superiority under ultra-low-dose conditions. The CTSpine1K dataset comprises $1,005$ CT volumes (over $500,000$ labeled slices and more than $11,000$ vertebrae), covering diverse morphological variations. Acquisition devices include scanners from GE, Philips, Siemens, and Toshiba. COVID-19 cases were selected for this study. Additionally, for the acquired mouse PCCT data, due to the complexity of the reconstruction algorithm, we also performed orthographic projection on the reconstructed images according to the provided reconstruction parameters. Similarly, the walnut data are publicly available dual-energy CT data with a projection size of $2063\times505$ and a reconstruction size of $1000\times1000$. Data from $15$ walnut samples were acquired using a customized miniature PCCT system. The energy thresholds were set to low-energy ($15-30$ $keV$) and high-energy ($30-80$ $keV$) bands. To simulate low-dose conditions, projection data were constructed based on the reference images obtained, thereby generating low-dose data for the study.
\subsection{Comparison of Advanced Self-Supervised Methods}
To fully validate the effectiveness of the proposed method, we selected eight self-supervised methods for comparison: BM3D \cite{dabov2006image}, B2U \cite{wang2022blind2unblind}, Neighbor2neighbor(NBR) \cite{huang2021neighbor2neighbor}, Noiser2noise \cite{moran2020noisier2noise}, PromptSID \cite{li2025prompt}, Noise2sim \cite{niu2022noise}, Noise2detail \cite{chobola2025lightweight}, and ZS-N2N \cite{mansour2023zero}. All methods were implemented according to their published papers. It is worth noting that most comparative studies typically use images of size $512\times512$. For the larger walnut dataset ($1000\times1000$), Neighbor2Neighbor struggled to process these dimensions. Therefore, we first downsampled the images to $512\times512$ for processing, then upsampled them back to $1000\times1000$. Additionally, for self-supervised methods requiring additional pre-training data, we trained them on the more formal Mayo2016 dataset before applying them to the mouse and walnut datasets for generalization.
\subsection{Evaluation Metrics}
To quantitatively evaluate image quality, this study follows the standard paradigm in medical imaging by treating reconstructed images from conventional doses as the reference baseline. This baseline is used to calculate three metrics: Peak Signal-to-Noise Ratio (PSNR), Structural Similarity Index (SSIM), and Root Mean Square Error (RMSE). Such high-dose images approximate the ideal noise-free state most closely, providing a reliable foundation for objective and reproducible comparative experiments. While not absolutely perfect, this remains the most practical and widely accepted gold standard for measuring imaging performance in the current stage. PSNR is defined as:

\begin{equation}
	PSNR=10\cdot\log_{10}\left(\frac{MAX_I^2}{MSE}\right)
	\label{eq8}
\end{equation}

where, $MAX_I$ represents the maximum pixel value that may appear in the image. SSIM is defined as:

\begin{equation}
	SSIM(x,y)=\frac{(2\mu_x\mu_y+C_1)(2\sigma_{xy}+C_2)}{(\mu_x^2+\mu_y^2+C_1)(\sigma_x^2+\sigma_y^2+C_2)}
	\label{eq9}
\end{equation}

among these, $\mu_x$ and $\mu_y$ represent the mean values within the selected window, $\sigma_x^2$ and $\sigma_y^2$ denote the variances, and $\sigma_{xy}$ signifies the covariance between the $x$ and $y$ components of the image, $C_1$ and $C_2$ are small constants employed to stabilize the division operation. RMSE is defined as:

\begin{equation}
	RMSE=\sqrt{\frac{1}{H\times W}\sum_{i=0}^{H-1}\sum_{j=0}^{W-1}[I_0(i,j)-I_0^{pre}(i,j)]^2}
	\label{eq10}
\end{equation}

where $I_0(\cdot)$ is the reference image, $I_0^{pre}(\cdot)$ is the denoised image.

\subsection*{Code availability}
Implementation code is available at \url{https://github.com/yqx7150/SCOUT}.


\subsection*{Acknowledgements}
Q. Liu is supported by the National Natural Science Foundation of China under Grants 621220033 and 62201193.
L. Shi is supported by the Nanchang University Youth Talent Training Innovation Fund Project under Grant XX202506030012, and the Early-Stage Young Scientific and Technological Talent Training Foundation of Jiangxi Province under Grant 20252BEJ730005.\\
M. Li is supported by the Jinan City Haiyou Talents Introduction Program.

\subsection*{Competing interests}
The authors declare no competing interests.

\clearpage

\bibliography{main}
\bibliographystyle{unsrt}

\clearpage


\end{document}